\documentclass[10pt,twocolumn,letterpaper]{article}

\usepackage{iccv}
\usepackage{times}
\usepackage{epsfig}
\usepackage{graphicx}
\usepackage{amsmath}
\usepackage{amssymb}
\usepackage{mathrsfs}
\usepackage[scr=boondox]{mathalfa}

\DeclareGraphicsExtensions{.pdf,.png,.jpg}
\DeclareMathOperator*{\argmax}{argmax} 
\usepackage{makecell}
\usepackage{booktabs}
\usepackage{multirow}
\usepackage{here}
\usepackage{kotex}
\usepackage{stackengine}

\usepackage[breaklinks=true,bookmarks=false]{hyperref}

\iccvfinalcopy 

\def\iccvPaperID{10492} 

\ificcvfinal\fi
\begin{document}
	
	\title{Self-supervised Learning of 3D Object Understanding by Data Association and Landmark Estimation for Image Sequence}
	
	\author{ 	
		Hyeonwoo Yu \qquad Jean On \\
		Robotics Institute, Carnegie Mellon University\\
		{\tt\small \{hyeonwoy,hyaejino\}@andrew.cmu.edu}
	}
	\maketitle
	
	\begin{abstract}
		In this paper, we propose a self-supervised learning method for multi-object pose estimation.
		3D object understanding from 2D image is a challenging task that infers additional dimension from reduced-dimensional information.
		In particular, the estimation of the 3D localization or orientation of an object requires precise reasoning, unlike other simple clustering tasks such as object classification.
		Therefore, the scale of the training dataset becomes more crucial.
		However, it is challenging to obtain large amount of 3D dataset since achieving 3D annotation is expensive and time-consuming.
		If the scale of the training dataset can be increased by involving the image sequence obtained from simple navigation, it is possible to overcome the scale limitation of the dataset and to have efficient adaptation to the new environment.
		However, when the self annotation is conducted on single image by the network itself, training performance of the network is bounded to the self performance.
		Therefore, we propose a strategy to exploit multiple observations of the object in the image sequence in order to surpass the self-performance:
		first, the landmarks for the global object map are estimated through network prediction and data association, and the corrected annotation for a single frame is obtained.
		Then, network fine-tuning is conducted including the dataset obtained by self-annotation, thereby exceeding the performance boundary of the network itself.
		The proposed method was evaluated on the KITTI driving scene dataset, and we demonstrate the performance improvement in the pose estimation of multi-object in 3D space.
	\end{abstract}
	
	\begin{figure}[t]
		\centering%
		\includegraphics[scale=0.09]{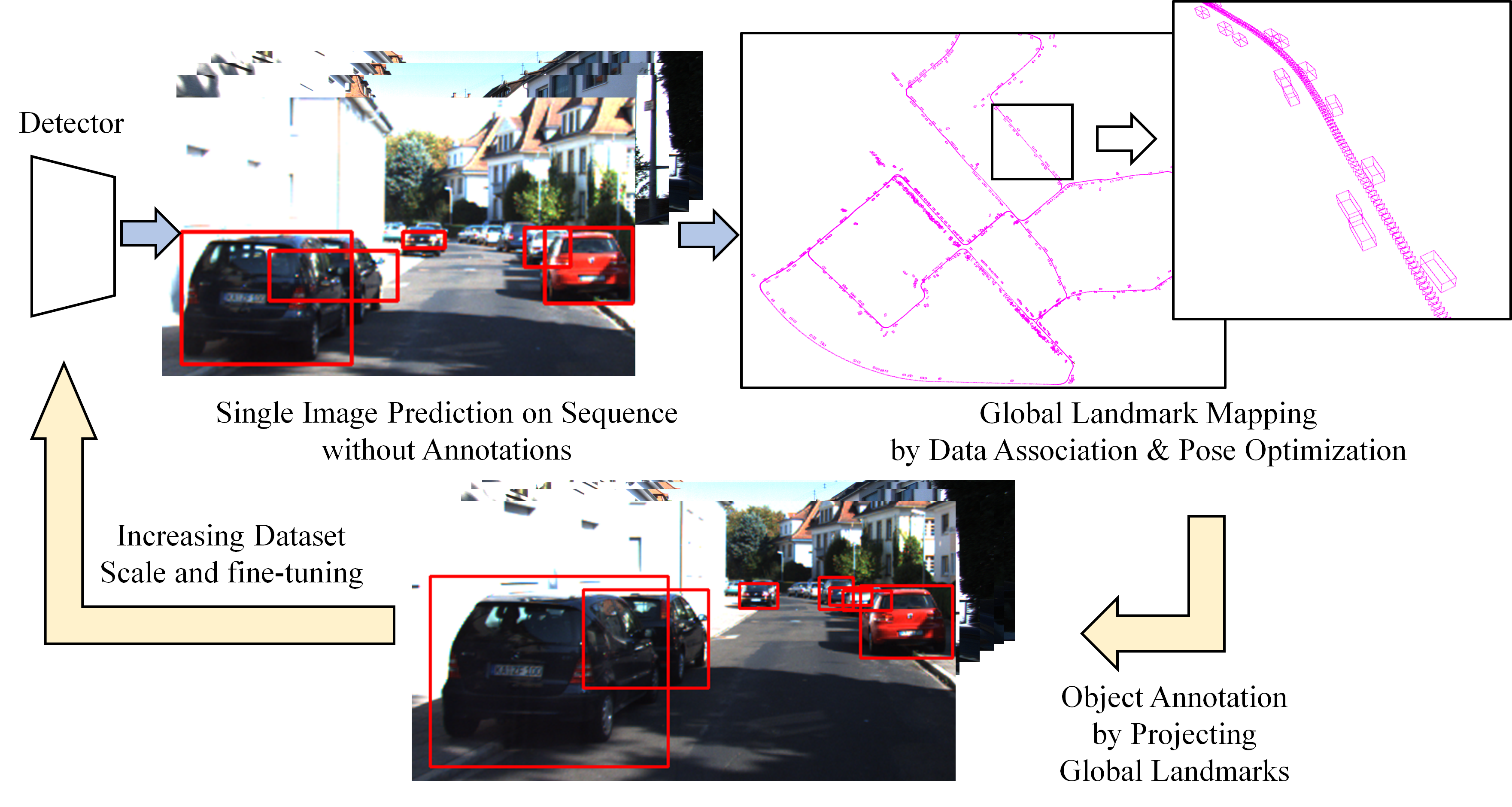}%
		\caption{%
			The overview of the proposed method.
			After trained on the original dataset, the network estimates the pose of multi-object for each image in sequence.
			Since the prediction results are inaccurate, we compensate errors by performing data association and pose estimation.
		}%
		\label{method_overview}%
	\end{figure}
	\section{Introduction}
	In the field of computer vision, understanding 3D objects is important for autonomous driving which is necessary for delivery drones, indoor robots and self-driving vehicles.
	Recently, a plethora of real-time visual perception techniques have been proposed \cite{fasterRCNN,yolo9000,bochkovskiy2020yolov4,retinanet,SSD} for semantic scene understanding or object-oriented simultaneous localization and mapping (SLAM) \cite{slam++, zhong2018detect}.
	In addition, beyond the simple object classification task, a number of techniques for extracting disentangled features have also been proposed to perform various tasks such as localization, orientation, or 3D shape reconstruction of objects \cite{M3DRPN,huo2020learning,tekin2018real,yu2018variational}.
	Unlike object classification, these techniques require accurate real-value inference, hence require sophisticated non-linear regression beyond simple clustering.
	In addition, the performance of the visual perception should not be highly dependent on the environment for the adaptive applications.
	To handle these issues, a complex network structure is required for elaborate regression using large amount of dataset \cite{simonyan2014very,huang2017densely}.
	In addition, in order to prevent overfitting and obtain robust regression results, a vast amount of training data suitable for network complexity is also required.
	
	Regression tasks in object understanding such as pose estimation require more datasets, because the output values can be any real values unlike objectness inference or classification.
	However, annotating on the dataset is a extremely expensive and time-consuming task that should be intervented by humans or human-level algorithms \cite{lin2014microsoft,xiang2014beyond,cordts2016cityscapes}.
	In addition, labeling 3D poses or shapes of objects or structures from single view data such as RGB or RGB-D is even more challenging.
	Even if a large amount of data is collected and labeled successfully, the need for training datasets for various environments persists, as the scope of the needs of autonomous vehicles or mobile drones are currently expanding globally.
	In addition, if home drones are commercially available, it is almost impossible to collect all training data in numerous indoor and outdoor environments for every home.
	In order to overcome these limitations, we can exploit the video data in the form of a sequencial dataset obtained while simple naviation or driving.
	For the image sequence acquired while driving, domain alignment can be performed preferentially before any labeling task.
	However, since this domain alignment is simply an alignment for a feature map at the network level, conditional alignment for a specific task is challenging.
	
	For the estimation of specific tasks such as disentangled feature estimation of an object, self-supervised learning should necessarily be performed.
	In particular, for the unlabled dataset we can perform rough labeling on the dataset using self annotation.
	However, if the network is exploited for the self annotation which will be used for its training again, the performance is bound to the network itself and scale of the original dataset used to train the  at the beginning.
	In other words, the obtained labeling result is bounded to the self-performance of the network, and even if training is performed using such labels, the performance cannot be significantly improved.
	To handle this limitation, it is possible to exploit multiple detection of objects existing in the video.
	Since the detection score depends on the position on the image or the distance from the camera, even the object is not properly observed in one frame, the observed value can be corrected through re-observation in a subsequent frame.
	In order to combine these multiple detection, the data association technique used in SLAM and the pose optimization of the landmark can be applied \cite{bowman2017probabilistic, yu2018variational, yang2019cubeslam}.
	Once the global landmark map is obtained for the sequence, robust annotations are obtained by projecting landmarks existing in the map for each image frame.
	By further learning the network using this new dataset, the weak point of prediction for a single image of the network is compensated, and at the same time, robust inference results for new environment are obtained.
	
	In this paper, self-supervised learning based on self annotation is applied to 3D object understanding of monocular image.
	The network infers depth and 3D orientation for object localization.
	We assume that the scale of the dataset with annotations is fixed, and the observed image sequence has no annotations and only camera poses at the time of observation are given.
	Through the proposed method, the network performs self annotation to acquire the dataset and conduct self-learning, but its performance is not self-bounded and shows more improved results.
	We verify the proposed technique for the KITTI dataset \cite{KITTIdataset} for the driving environment.
	As a dataset including annotations, the KITTI object 3D detection dataset is used.
	As a non-annotated dataset, the KITTI odometry dataset, which is an image sequence, is used.
	When self-supervised learning is performed using the proposed technique, performance improvement is confirmed in the pose estimation of the object, especially orientation estimation.
	\begin{figure*}[t]
		\centering%
		\includegraphics[scale=0.33]{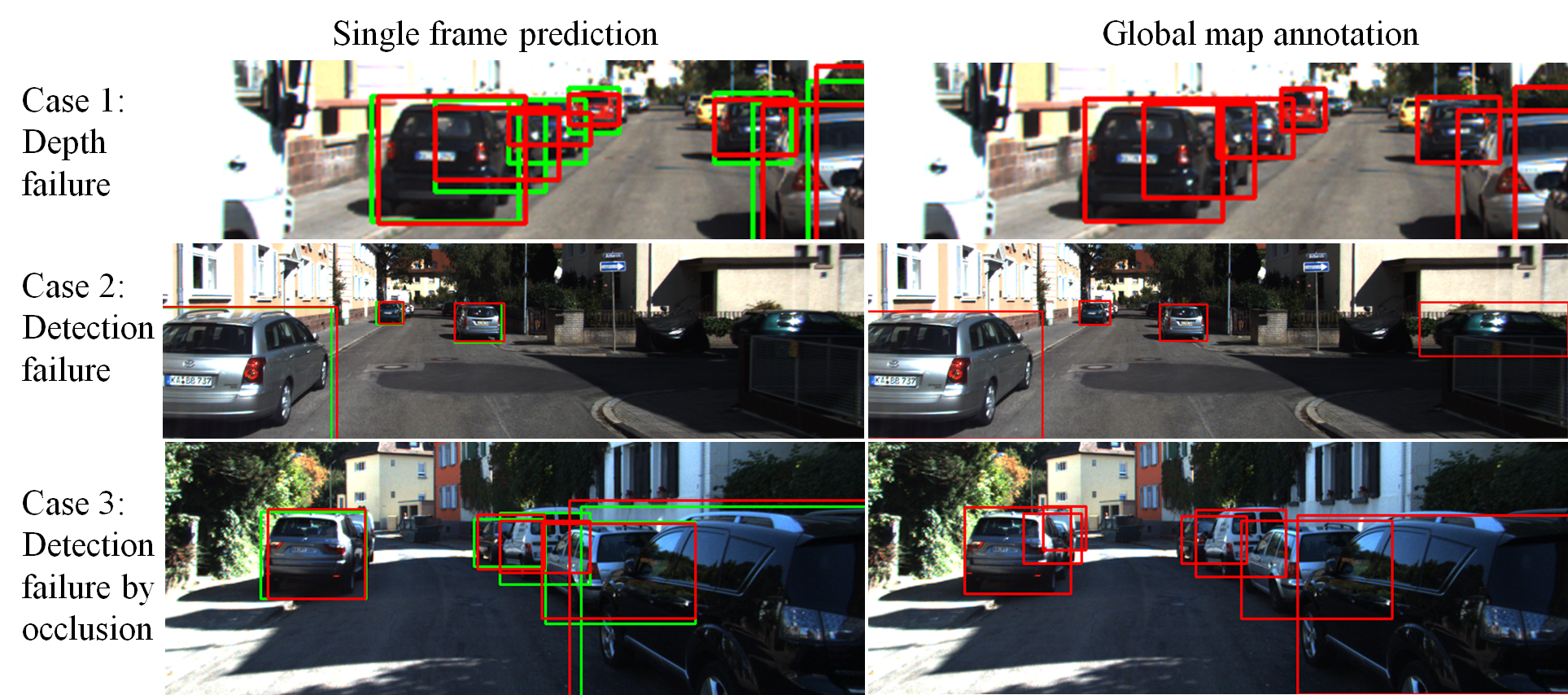}%
		\caption{%
			Examples of the failure cases of single-image estimation and their corrections using global map.
			The red box and the green is the projected 3D bounding box and 2D bounding box, respectively.
			For comparison, we show the green 2D bounding boxes obtained by other detection method.
			As shown in left side, in single-frame detection depth estimation are not accurate (top-left). Also some objects are not detected due to the threshold (middle-left) or occlusion (bottom-left).
			These cases can be corrected by projecting global landmarks in global map as shown in right side.
		}%
		\label{failure_example}%
	\end{figure*}

	\section{Related Works}
	Due to the nature of deep learning based on statistics, the scale of training data has a great influence on the performance of the network.
	Therefore, in order to obtain robust estimation in various environments, a large-scale training dataset is needed for fully utilizing the network capacity without overfitting.
	With various 3D sensors such as LIDAR and RGB-D camera and inexpensive 2D cameras, data collection becomes fast and efficient.
	However, obtaining high-quality labeling still takes a lot of time and effort.
	In particular, the annotation for real-value features of object understanding such as localization or pose estimation requires human-level intervention.
	We introduce the related works to overcome the limitation as follows.
	\subsection{Domain Alignment}
	With the advent of the neural network, deep features can be easily obtained in the unsupervised manner such as Autoencoder. In addition, encoders that can extract those deep features can effectively be trained through SGD can be freely fine-tuned with any constraint.
	Therefore, a number of studies have been proposed to reliably cope with the unlabeled dataset by performing domain alignment at the feature level \cite{saenko2010adapting,hoffman2013efficient,moon2020multi,ganin2016domain,chen2018domain}.
	Through these methods, it is achievable to perform more robust category classification or clustering for other domains, or to further improve the performance of determining objectness for multi-object scenes.
	However, since these approaches eventually aligns the feature map without manual annotation, it cannot be guaranteed to be conditional aligned to a specific task \cite{sun2016return,gretton2012kernel}.
	In other words, an inevitable error appears in elaborate real value inference such as the pose of an, due to the limitation of the unconditioned unsupervised learning.
	
	\subsection{Self-supervised Learning for Object Detection}
	The most effective in practical network learning is to directly learning the likelihood with the assigned labels for specific tasks.
	Therefore, in order to overcome the limitations of the data scale, various self-supervised learning techniques have been proposed.
	In particular, techniques for multi-object detection in 2D images have been proposed \cite{amrani2020self,li2020improving,liu2020self,NEURIPS2019_d0f4dae8,bansal2018zero}.
	However, these techniques have limitations in applying to the 3d pose of an object, considering only projected features, such as a bounding box on a 2D image.
	A number of studies have been proposed that apply self-supervised learning to 3D object detection using spatial sensor information such as RGB-D images \cite{dai2020sg,zhao2020sess}.
	These techniques properly utilize 3D information to efficiently perform 3D detection even for an unknown scene.
	However, it is difficult to apply to challenging situations that are dimension-reduced when only 2D images are used.
	In addition, all of the above techniques lack of the consideration for the situation of using a sequence dataset that can be easily obtained while navigation.
	
	\subsection{Data Association and Pose Optimization}
	In the sequence dataset, it is the huge advantage that the same object is detected repeatedly in multiple frames.
	Therefore, even if there are erroneous observations for an object, the observation results can be compensated by associating multiple observations.
	However, when the camera moves, the position and scale of the object on the 2D image plane both varies. Therefore, data association that can classify it as the same object must be performed.
	Landmarks existing on 3D space are acquired by data association.
	The corrected pose for this landmark are obtained by classifying the outlier for the observation result and obtaining the average pose.
	This series of processes is essentially used in most SLAM techniques using sequence input \cite{slam++,categorySpecificSLAM,bowman2017probabilistic,yu2018variational,zhong2018detect, yu2019variational, frost2016object, yang2019cubeslam, yang2019monocular}.
	Therefore, in this paper, a global landmark map is obtained by applying the data association technique used in the various SLAM methods.
	After that, by projecting the landmarks in reverse to each image, a robust annotation result using the landmark with the corrected pose is obtained.
	The network corrects detection failure by using the best detection results, and can go beyond the boundary of self-performance.
	In the proposed technique, camera pose estimation is not performed, and it is assumed that a camera pose is given for a sequence.
	The technique most similar to the proposed technique is \cite{zhong2018detect}.
	While performing SLAM using RGB-D, if the moment when tracking fails occurs, labeling is performed using previous observation information.
	The 2D object detection network is reinforced using the acquired labeling.
	However, this technique is limited to 2D detection, and 3D localization and pose estimation of the object are not considered.

	\section{Approach}
	In terms of inferring additional dimensional information such as estimating 3D perception from a monocular image, requires a lot of training data.
	However, 3D annotation for 2D image dataset needs too much effort and also time-consuming. It also requires direct human-level intervention.
	For this case, self annotation can be performed using the trained network with the limited training dataset.
	With self-annotation approach, a new training dataset can be additionally created for a new image dataset from a new environment.
	However, it is still challenging to obtain high-quality annotations because the network with limited performance bounded to the original dataset is exploited.
	Furthermore, in the case of multi-object understanding, a false positive or true negative for objectness estimation becomes a problem.
	For example, suppose that depth and orientation of multi-object are estimated from a single image.
	Then, the following limitations exist;
	first, localization may fail due to incorrect depth estimation.
	Next, pose estimation failure can be occurred due to the incorrect viewpoint prediction.
	To eliminate outliers like these two cases, the confidence threshold for the objectness can be set as high value.
	However, this simple strategy can reduce the recall of detecting and it is going to be hard to achieve the robust annotation.
	
	Even if the self annotation is performed successfully on a single image, the obtained labels are bounded by the self performance of the network.
	For example, let $\{\mathcal{X}_s, \mathcal{L}_s\}$ be the training dataset.
	In our case, $\mathcal{X}_s$ and $\mathcal{L}_s$ are the image set and the label set, respectively.
	Also, let $\mathcal{X}_t$ be the dataset without any annotations.
	Then the annotation set $\mathcal{L}_t$ according to $\mathcal{X}_t$ can be estimated as the following:
	\begin{align}
	\label{optimal_l}
		\mathcal{L}_t^* = \argmax_{\mathcal{L}} p_\theta \left(
															\mathcal{L} | \mathcal{X}_t; \mathcal{X}_s, \mathcal{L}_s
														\right),
	\end{align}
	where $\theta$ is the network parameter.
	By using $\mathcal{L}_t^*$ as the self-annotation set, we can perform estimations for the arbitrary dataset $\mathcal{X}$ as:
	\begin{align}
	\label{optimal_l_arbitrary}
	\mathcal{L}^* = \argmax_{\mathcal{L}} p_\theta \left(
								\mathcal{L} | \mathcal{X}; \mathcal{X}_s, \mathcal{L}_s, \mathcal{X}_t, \mathcal{L}_t^*
								\right)
	\end{align}
	Here, $\mathcal{L}_t^*$ is obtained by \eqref{optimal_l} so that it is dependent to $\mathcal{X}_s, \mathcal{L}_s$ and $\theta$.
	Therefore we can rewrite \eqref{optimal_l_arbitrary} as the following:
	\begin{align}
	\label{optimal_l_arbitrary_final}
	\mathcal{L}^* = \argmax_{\mathcal{L}} p_\theta \left(
							  \mathcal{L} | \mathcal{X}; \mathcal{X}_s, \mathcal{L}_s
							\right)
	\end{align}
	Consequently, \eqref{optimal_l_arbitrary_final} has no difference compared to \eqref{optimal_l} and the result $\mathcal{L}^*$ is bounded to the original dataset $\{\mathcal{X}_s, \mathcal{L}_s\}$ under $\theta$.
	As a specific example, assume that the network is trained to estimate depth $z$ of the object from single 2D image.
	Then the euclidian loss function of $z$ can be defined as:
	\begin{align}
	\label{depth_loss}
	e = \lVert z_{label} - z_{pred} \rVert^2
	\end{align}
	Here, suppose $z_{label}$ is obtained by self annotation using \eqref{optimal_l_arbitrary_final}.
	Then $z_{label}$ is eventually equivalent to a prediction result $z_{pred}$ from the trained network, therefore the error $e$ becomes constant and the network is not going to be updated.
	In other words, even though self annotation is performed, a dataset bounded by the existing training dataset is obtained.	

	To overcome this problem, an image sequence in the form of the video can be exploited.
	Suppose that the camera pose for each image is given.
	Then multiple observations of each object can be obtained from the sequence.
	Therefore, even the observations are obtained from single network, observations can be corrected by taking average of the multiple observations.
	By projecting this average pose observations to each image plane, we can obtain smoothed annotations for the entire sequences.
	In this way, it is possible to reliably correct poor detection results for objects that are not constantly observed in image sequence.
	Also, by obtaining a new training dataset including the corrected annotation for a new dataset, the network can learn about a new environment and achieve high performance.
	In order to verify the proposed method, we set depth and orientation estimation as main task, which is essential for object pose estimation in monocular image.
	We display the examples of limitations of the single-image detection and their successful corrections in Fig.~\ref{failure_example}.
	\begin{figure*}[t]
		\centering%
		\includegraphics[scale=0.22]{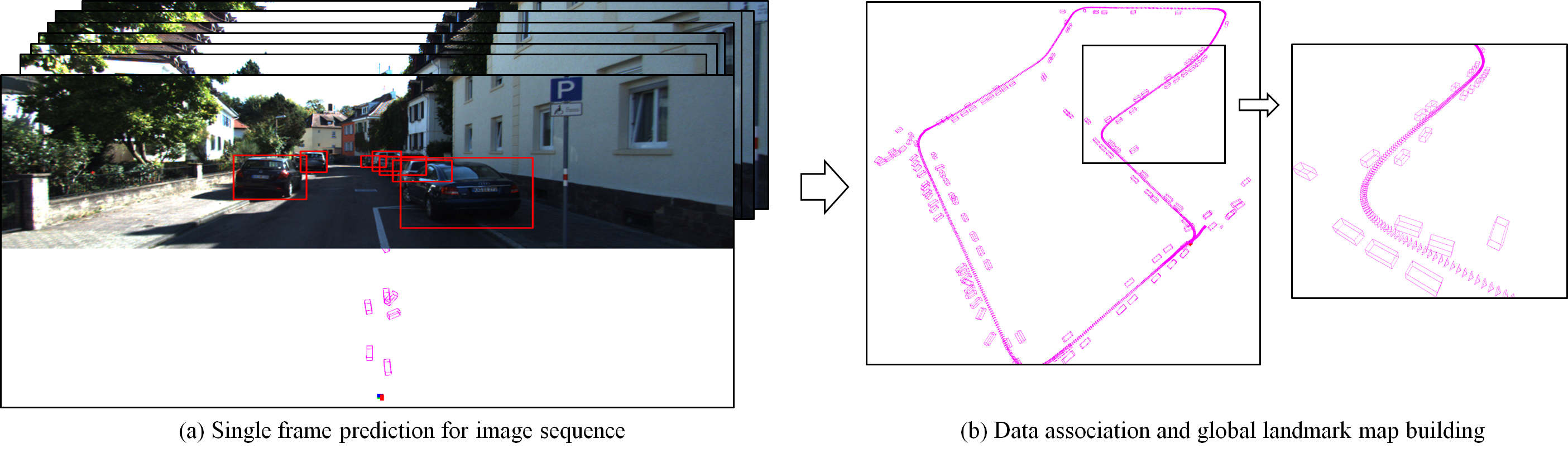}%
		\caption{%
			The overview of the global landmark map estimation.
			First, the network estimates the depth and orientation of multi-object from single image.
			In this stage the prediction results have detection noise.
			By performing data association and pose estimation for entire image sequence, we can remove outliers and reduce the noise of the predictions.
			The global map involves static landmarks with smoothed pose by considering objectness or pose uncertainty.
		}%
		\label{global_map_example}%
	\end{figure*}
	\subsection{Network Prediction for Single Image}
	By using the network trained until fully converged on the original training dataset, we obtain the prediction results on the sequence dataset for self annotation.
	In this case the recall changes drastically according to the confidence threshold of the objectness.
	Since the proposed technique should not be bound by self performance, we set the threshold high enough so that the precision is very high even if the recall is small.
	In the case of single shot prediction, due to this setting a lot of multi-object can be missing.
	For example, an occluded object or an object that is too far away from the camera will not be detected with low confidence threshold.
	However, even an object that has not been detected in the current frame can be detected in a subsequent frame as the image sequence is exploited.
	Therefore, in the aspect of the entire frame, precise detection can be obtained while maintaining high recall;
	once the global landmark map is given, more sophisticated annotations can be obtained with this approach.

	\subsection{Data Association}
	In order to track the multiple observations of multi-object for an image sequence, data association is essential.
	For the data association, several object detection features can be exploited:
	first, since we use the multi-object detector in a 2D image plane, 2D bounding boxes can be compared by calculating 2D intersection over union (IoU) \cite{wu2020eao}.
	For 2D bounding box, 3D bounding box estimated from the object's pose can be projected to the 2D image plane as the following:
	\begin{align}
	\label{3d_2d_projection}
	\begin{bmatrix}
	x_{2D}  \cdot z_{3D} \\ y_{2D} \cdot z_{3D} \\ z_{3D}
	\end{bmatrix}
	=
	\boldsymbol{P} \cdot
	\begin{bmatrix}
	x_{3D} \\ y_{3D} \\ z_{3D} \\ 1
	\end{bmatrix},
	\end{align}
	where $\left(x_{3D}, y_{3D}, z_{3D}\right)$ is the 3D coordinate of the 3D point, $\left(x_{2D}, y_{2D}\right)$ is the 2D pixel coordinate according to the 3D point and $\boldsymbol{P}$ is the projection matrix.
	Then, after calculating the 2D bounding box for each image of the sequence, the association of the object can be determined by calculating the IoU of the 2D bounding box between the two images.
	However, when the rotation angle of the camera trajectory is too large or the speed of the camera movement is too fast, it is not enough to solely consider the 2D IoU for data association.
	To overcome this aspect, visual features can be additionally used.
	By performing feature matching using SIFT or SURF, we can perform object tracking when a feature matching rate is above a specific threshold \cite{yang2019cubeslam, yang2019monocular}.
	However, data association may fail if features are not created enough.
	For example, when the object is too small in image plane or the object is far away from the camera, or the edge is insufficient, these situation occurs textureless observations.
	In addition, when an object is occluded by other objects or obstacles, it is challenging to perform accurate matching as some parts of other objects exist in the 2D bounding box.
	Also, these two approaches are not suitable when there is a gap between the first observed frame and the later observed frame.
	Therefore, to compensate for this limits, we can compare the locations of objects in the global 3D space.
	However, in this case, since localization depends entirely on the detector network itself, it may lead to incorrect data association results depending on the estimation error.
	Therefore, in the proposed method, we perform data association by using all three methods.
	
	\subsection{Global Map Building}
	We can obtain global landmark map using multiple observations for static objects.
	Each observation represents a pose relative to the camera pose at the observed time point.
	Therefore, when the camera pose at the time of observation is given, the object pose can be converted into a global pose as follows:
	\begin{align}
	\label{local_to_global}
	X_0 = C^{k}_0 \cdot X_{k},
	\end{align}
	where $X_{k}$ is the object pose related to the camera pose at $k$th frame. $X_0$ and $C^{k}_0$ are poses of the object and camera in global coordinate, respectively.
	Since the detector also estimates the detection score for the observation of the object, each multiple observation involves a detection score.
	Therefore, we can weight more accurate observations and discard uncertain results by taking a weighted average using this score.
	Then, the average of the object's depth and orientation can be obtained as follows:
	\begin{align}
	\label{z_average}
	z_{avr} &= \frac{\sum w_i \cdot z_i }
	{\sum w_i}
	\\
	\nonumber
	M &= \frac{\sum w_i \cdot \boldsymbol{R}_i}
	{\sum w_i}
	=  U^* \cdot D^* \cdot V^t
	\\
	\label{R_average}
	\boldsymbol{R}_{avr} &= U^* \cdot V^t
	,
	\end{align}
	where $\boldsymbol{R}$ is the rotation matrix of the object orientation.
	Here, we assume that the orientation of the object only has a yaw rotation.
	Different to the object, since the camera pose have 3 degree of freedom (roll, pitch and yaw), after computing object's pose by \eqref{local_to_global} we can recompute the yaw angle of the object by the following:
	\begin{align}
	\label{ry_estimation}
	r_y = arctan2\left(
	-R_{31}
	,
	\left(R_{11}^2 + R_{21}^2\right)^{\frac{1}{2}}
	\right)
	\end{align}
	Using \eqref{z_average}, \eqref{R_average} and \eqref{ry_estimation}, we can get the optimized poses for landmarks in the global 3d space.
	An example of the obtained global object map is shown in Fig.~\ref{global_map_example}.
	
	\subsection{2D Projection}
	Object annotation can be obtained for each image of sequence by using the global landmark map.
	Given a camera pose in an arbitrary $k$th frame, the local pose of the object can be obtained as:
	\begin{align}
	\label{global_to_local}
	X_k = \left(C^{k}_0\right)^{-1} \cdot X_{0}
	\end{align}
	Using the object's local pose and dimensions of 3D bounding box, we can get 8 corner points of 3D bounding box.
	Then, the center and 8 points of the object can be projected onto the 2D image plane using \eqref{3d_2d_projection}.
	Here, if the depth of the object is less than $0$, that is, if the object exists behind the 2D image plane, it cannot be observed in the current image, therefore it is excluded from the current frame.

	With this global map projection, all objects detected before and after the current frame can also be annotated.
	In practice, detection often fails or inference fails even if the same object exists in a continuous image sequence.
	In this case, by using the global landmark map projection, a certain object annotation for each frame is possible.
	In other words, even if the object is not detected by the network in a certain frame, annotation of the object can be achieved in that frame.
	In order to discard objects observed in frames that are too future from the current frame, the frame threshold is also applied in actual implementation.
	
	\begin{table*}[t]		
		\begin{center}
			\begin{tabular}{c | c |c  c c c c c}
				\hline
				\hline
				\multirow{2}{*}{Method}  & \# of   & $\sigma < {1.25}$ &  & Abs Rel & Sqr Rel & RMSE & RMSE$_{log}$ \\
				& predictors & (higher is better) && \multicolumn{4}{c}{(lower is better)} \\
				\hline
				SVR\cite{gokcce2015vision} & - & 0.345 &  & 1.494 & 47.748 & 18.970 & 1.494 \\
				IPM\cite{tuohy2010distance} & - & 0.701 &  & 0.497 & 35.924 & 15.415 & 0.451 \\
				\hline
				Zhu et al.\cite{zhu2019learning} & - & 0.848 & & 0.161 & 0.619 & 3.580 & 0.228 \\
				\hline
				Zhang et al.\cite{zhang2020regional} & - & \textbf{0.992} &  & \textbf{0.049} & - & \textbf{1.931} & - \\
				\hline
				DORN \cite{fu2018deep}  & - & 0.883 & & 0.190 & 1.153 & 4.802 & 0.287 \\
				\hline\hline										  
				\multirow{1}{*}{Yu et al. \cite{yu2021anchor}}
				& 9 & 0.970 & & 0.079 & 0.165 & 1.719 & 0.127 \\
				\cline{1-1}\cline{2-8}
				\multirow{1}{*}{baseline}
				& \multirow{3}{*}{12} & 0.9962 / 0.9937 & & 0.0419 / 0.0472 & 0.0593 / 0.0662 & \textbf{1.2965} / 1.2602 & 0.0554 / 0.0633 \\			
				\cline{1-1}\cline{3-8}										  
				\multirow{1}{*}{proposed}
				&  & 0.9957 / 0.9936 & & \textbf{0.0408} / \textbf{0.0450} & \textbf{0.0584} / \textbf{0.0609} & 1.3009 / \textbf{1.2256} & \textbf{0.0549} / \textbf{0.0610} \\
				\cline{1-1}\cline{3-8}										  
				\multirow{1}{*}{\textit{upper bound}}
				&  & 0.9954 / 0.9941 & & 0.0399 / 0.0424 & 0.0552 / 0.0564 & 1.2445 / 1.1899 & 0.0550 / 0.0586 \\
				\hline\hline				
			\end{tabular}
		\end{center}
		\caption{Comparison of the depth estimation for KITTI dataset (split1/split2)}
		\label{depth_eval}
	\end{table*}
	\begin{table*}[h]	
		\begin{center}
			\begin{tabular}{c | c c c c }
				\hline
				\hline
				& \textit{Acc}$_{\frac{\pi}{4}}$ (\%) & \textit{Acc}$_{\frac{\pi}{6}}$ (\%) && \textit{MedErr} (degree)  \\
				& \multicolumn{2}{c}{(higher is better)} && (lower is better) \\
				\hline
				baseline & 0.768 / 0.716 & 0.716 / 0.649 && 10.737 / 12.994 \\
				\hline
				proposed & \textbf{0.779} / \textbf{0.721} & \textbf{0.729} / \textbf{0.671} && \textbf{8.270} / \textbf{8.532} \\
				\hline
				\textit{upper bound} & 0.796 / 0.721 & 0.743 / 0.677 && 8.354 / 9.462 \\
				\hline
			\end{tabular}
		\end{center}
		\caption{Comparison of the Viewpoint Estimations on KITTI dataset (split1/split2)}
		\label{viewpoint_eval}
	\end{table*}
	\begin{table*}[t]		
		\begin{center}
			\begin{tabular}{c | c | c c c c}
				\hline
				\hline
				(\textit{unit : meter}) & & Abs Rel & Sqr Rel & RMSE & RMSE$_{log}$ \\
				\hline
				\multirow{2}{*}{0$\sim$10}
				& baseline & 0.0582 / \textbf{0.0743} & 0.0439 / 0.0699 & 0.5523 / 0.6946 & 0.0799 / \textbf{0.0971} \\
				& proposed & \textbf{0.0582} / 0.0758 & \textbf{0.0411} / \textbf{0.0686} & \textbf{0.5447} / \textbf{0.6753} & \textbf{0.0783} / 0.0988 \\
				\hline
				\multirow{2}{*}{10$\sim$20}
				& baseline & 0.0498 / 0.0572 & 0.0625 / 0.0779 & 0.9752 / 1.0775 & 0.0668 / 0.0752 \\
				& proposed & \textbf{0.0470} / \textbf{0.0521} & \textbf{0.0597} / \textbf{0.0652} & \textbf{0.9564} / \textbf{0.9855} & \textbf{0.0665} / \textbf{0.0691} \\
				\hline
				\multirow{2}{*}{20$\sim$30}
				& baseline & 0.0424 / 0.0439 & \textbf{0.0628} / 0.0678 & \textbf{1.2310} / 1.2781 & 0.0517 / 0.0539 \\
				& proposed & \textbf{0.0412} / \textbf{0.0409} & 0.0633 / \textbf{0.0623} & 1.2446 / \textbf{1.2298} & \textbf{0.0516} / \textbf{0.0513} \\
				\hline
				\multirow{2}{*}{30$\sim$40}
				& baseline & 0.0339 / 0.0326 & 0.0554 / 0.0501 & 1.3733 / 1.3100 & 0.0407 / 0.0386 \\
				& proposed & \textbf{0.0330} / \textbf{0.0316} & \textbf{0.0535} / \textbf{0.0486} & \textbf{1.3519} / \textbf{1.2889} & \textbf{0.0399} / \textbf{0.0381} \\
				\hline
				\multirow{2}{*}{40$\sim$50}
				& baseline & \textbf{0.0305} / \textbf{0.0281} & 0.0552 / 0.0482 & \textbf{1.5573} / \textbf{1.4512} & \textbf{0.0356} / \textbf{0.0334} \\
				& proposed & 0.0306 / 0.0284 & 0.0562 / \textbf{0.0492} & 1.5721 / 1.4657 & 0.0359 / 0.0337 \\
				\hline
				\multirow{2}{*}{50$\sim$}
				& baseline & \textbf{0.0298} / \textbf{0.0299} & \textbf{0.0719} / \textbf{0.0705} & \textbf{2.0864} / \textbf{2.0492} & \textbf{0.0350} / \textbf{0.0349} \\
				& proposed & 0.0302 / 0.0300 & 0.0749 / 0.0721 & 2.1307 / 2.0723 & 0.0358 / 0.0354 \\
				\hline
				\hline				
			\end{tabular}
		\end{center}
		\caption{Comparison of the Interval Depth Estimations on KITTI dataset (split1/split2)}
		\label{depth_eval_interval}
	\end{table*}
	\begin{table*}[t]		
		\begin{center}
			\begin{tabular}{c | c | c c c }
				\hline
				\hline
				(\textit{unit : meter}) & & \textit{Acc}$_{\frac{\pi}{4}}$ & \textit{Acc}$_{\frac{\pi}{6}}$ & \textit{MedErr} \\
				\hline
				\multirow{3}{*}{0$\sim$10}
				& baseline & \textbf{0.9255} / \textbf{0.8554}  & \textbf{0.8989} / 0.7798  & 4.5797 / 6.1599 \\
				& proposed (w/o uncertainty) & 0.8906 / 0.8088  & 0.8592 / 0.7782  & \textbf{4.3056} / 4.3910 \\
				& proposed (w uncertainty) & \quad\,--\,\quad / 0.8183 & \quad\,--\,\quad / \textbf{0.7902} & \quad\,--\,\quad / \textbf{4.3695} \\
				\hline
				\multirow{3}{*}{10$\sim$20}
				& baseline & \textbf{0.8464} / \textbf{0.7555}  & \textbf{0.8094} / \textbf{0.6869}  & 7.6984 / 10.6778 \\
				& proposed (w/o uncertainty) & 0.8062 / 0.6878  & 0.7651 / 0.6367  & \textbf{6.4741} / 8.0451 \\
				& proposed (w uncertainty) & \quad\,--\,\quad / 0.6737 & \quad\,--\,\quad / 0.6340 & \quad\,--\,\quad / \textbf{7.7723} \\
				\hline
				\multirow{2}{*}{20$\sim$30}
				& baseline & 0.7883 / 0.7007  & 0.7204 / 0.6291  & 12.9076 / 13.6581 \\
				& proposed (w/o uncertainty) & \textbf{0.7924} / \textbf{0.7205}  & \textbf{0.7358} / \textbf{0.6719}  & \textbf{7.8522} / 8.3518 \\
				& proposed (w uncertainty) & \quad\,--\,\quad / 0.7137 & \quad\,--\,\quad / 0.6685 & \quad\,--\,\quad / \textbf{7.6802} \\
				\hline
				\multirow{3}{*}{30$\sim$40}
				& baseline & 0.7490 / 0.6800  & 0.6733 / 0.6141  & 13.3858 / 15.6678 \\
				& proposed (w/o uncertainty) & \textbf{0.8008} / 0.7564  & \textbf{0.7562} / 0.7109  & \textbf{9.0621} / 8.2972 \\
				& proposed (w uncertainty) & \quad\,--\,\quad / \textbf{0.7626} & \quad\,--\,\quad / \textbf{0.7195} & \quad\,--\,\quad / \textbf{7.8145} \\
				\hline
				\multirow{3}{*}{40$\sim$50}
				& baseline & 0.7004 / 0.6424  & 0.6526 / 0.5817  & 12.7294 / 18.4677 \\
				& proposed (w/o uncertainty) & \textbf{0.7913} / 0.7191  & \textbf{0.7379} / 0.6654  & \textbf{9.3358} / 13.4758 \\
				& proposed (w uncertainty) & \quad\,--\,\quad / \textbf{0.7201} & \quad\,--\,\quad / \textbf{0.6687} & \quad\,--\,\quad / \textbf{12.7727} \\
				\hline
				\multirow{3}{*}{50$\sim$}
				& baseline & 0.4482 / 0.5417  & 0.3956 / 0.4983  & 59.0119 / 30.3138 \\
				& proposed (w/o uncertainty) & \textbf{0.4908} / \textbf{0.6096}  & \textbf{0.4023} / \textbf{0.5284}  & \textbf{46.2874} / \textbf{27.6205} \\
				& proposed (w uncertainty) & \quad\,--\,\quad / 0.5965 & \quad\,--\,\quad / 0.5122 & \quad\,--\,\quad / 28.3952 \\
				\hline				
			\end{tabular}
		\end{center}
		\caption{Comparison of the interval Viewpoint Estimations on KITTI dataset (split1/split2)}
		\label{viewpoint_eval_interval}
	\end{table*}
	
	\subsection{Data Uncertainty}
	When creating a new dataset by self labeling using a global landmark map, the score (or detection confidence) can be used as weights as in \eqref{z_average} and \eqref{R_average};
	the impact of uncertain detection results can be reduced, and the impact of certain results can be increased.
	However, in this case, since confidence of \textit{objectness} is reflected, it is hard to reflect uncertainty in pose estimation.
	Meanwhile, by inferring the uncertainty and pose together, we can use this uncertainty as a weight instead of the score of objectness.
	In the case of inferring object's depth $z$, loss $e$ including the uncertainty $\sigma$ can be expressed as the following \cite{kendall2017uncertainties}:
	\begin{align}
	\label{depth_loss_uncertainty}
		e = \frac{1}{2\sigma^2} \lVert z_{label} - z_{pred} \rVert^2
		    + \frac{1}{2}\log \sigma^2
	\end{align}
	By using $\sigma$ obtained using \eqref{depth_loss_uncertainty} instead of \eqref{depth_loss} as a weight, we can perform pose optimization by considering the confidence for pose estimation, not the confidence of objectness.

	\section{Experiments}
	\subsection{Network}
	To verify the robustness of the proposed method and represent the upper bound of self annotation, We use a network capable of real-time performance based on the YOLOv2\cite{yolo9000} structure as a baseline \cite{yu2021anchor}.
	Unlike the original network structure, predictors are set to $12$ per grid.
	In all experiments, self annotation is performed on the image sequence after the network training converges on the original training dataset.
	The obtained sequence dataset is then used for fine-tuning the network.
	We also perform additional experiments to check the upper bound of self-supervised learning based on the proposed self annotation.
	However, in this case, it is necessary to have the ground truth of the object detection annotaion for the sequence dataset.
	To replace this, the same self-annotation process is performed using the 3D detection network \cite{M3DRPN} with better detection performance but a lower real-time performance than the baseline network (DenseNet-121 \cite{huang2017densely} vs Darknet19 \cite{yolo9000} as backbone).
	The obtained annotations are treated as ground truth annotations, and the upper bound of the proposed method is evaluated by training the baseline network on this annotations.

	\subsection{Dataset and Training}
	In this paper, we set multi-object understanding in 2D image as the main task.
	As a dataset with ground truth annotation, KITTI object 3D detection \cite{KITTIdataset} dataset is used.
	The main purpose of the network is to infer the depth ($z$-axis distance) and the orientation of an object.
	The orientation is defined as the orientation on the image in order to reduce the burden on the network \cite{M3DRPN, huo2020learning}.
	In addition, we assume that the object exists on the ground plane and only yaw angle ($y$-axis) rotation exists.
	We conduct all experiments on two common data splits including val1 \cite{chen20153d} and val2 \cite{xiang2017subcategory}.
	
	We use \texttt{Car} as the object category.
	The KITTI odometry dataset \cite{KITTIdataset} is used as a sequencial dataset for self-annotation.
	In this dataset, there is no annotation on object detection, and only the camera pose and projection matrix for each frame are given.
	Among a total of 11 sequences, sequences in which a large number of dynamic objects exist are excluded, such as highway scenes.
	In addition, sections in which too many overlapping objects exist, such as vehicles continuously parked within a sequence, or scenes in which dynamic objects appear are also excluded.
	
	The size of the input image of the network is fixed to $1216\times486$.
	During network training, common data augmentation such as Gaussian noise, hue sat or RGB conversion is not applied to the image.
	The vertical flipping is only adopted for the augmentation.
	For the solver for gradient update we use Adam with the learning rate 1e-4 and batch size 8 in all cases.
	After converting the sequence dataset to the new training dataset with self annotation, training is conducted in the same way for the original dataset.
	
	\subsection{Evaluation}
	To verify the effect of the dataset scale, we evaluate the proposed method for the object's depth and orientation.
    By using the anchor distance, the baseline network is suitable for depth estimation \cite{yu2021anchor}.
    In addition, we tend to achieve more precise depth estimation by increasing the number of predictors (anchor distance) to 12.
    We verify how effective the increase of the dataset scale by self annotation even in this setting.
    Compared to the depth, anchor for orientation are not considered.
    Therefore, each predictor still has to infer the rotation of all ranges with the dicreased number of training datapoints for each predictor.
    We also verify the effectiveness of self-supervised learning in this challenging situation in contrast to the depth estimation.
	
	We use the evaluation metric used in \cite{zhu2019learning,zhang2020regional} for depth estimation of the object.
	For orientation, we measure \textit{Acc}$_{\frac{\pi}{4}}$, \textit{Acc}$_{\frac{\pi}{6}}$ and \textit{MedErr} \cite{tulsiani2015viewpoints,mousavian20173d}.
	
	\subsubsection{Distance}
	We represent the depth estimation results in Table.~\ref{depth_eval}.
	Since our baseline uses more anchor distances than \cite{yu2021anchor}, it shows high performance in depth estimation compared to the traditional method.
	Even in this setting, performance is improved when fine tuning of the network on the self-annotated dataset.
	Also, for all splits, we found that there is a slight performance improvement in almost all evaluation metrics.
	Since the distance range covered by a predictor is within 5m, the improvement in depth estimation performance seems to be not drastic.
	
	Table.~\ref{depth_eval_interval} shows the estimation results according to the depth interval.
	In the range $0\sim40m$, we figure out the evident performance improvement.
	However, we also found that objects in 40 or more have similar performance or have poor performance compared to the baseline.
	This is because when performing self annotation by projecting an image from a global landmark map, an object observed in a future frame too far from the current frame was not used for labeling.
	As a result, several objects at quit far distance (or frame) from the current camera pose were not included in the dataset.
	If occlusion can be considered by considering other features such as segmentation, it is possible to determine whether a distant object exists in the current frame and include them in the training dataset.
	
	\subsubsection{Rotation}
	We represent the evaluation results of the orientation in Table.~\ref{viewpoint_eval}.
	Since the network used as the baseline has 12 predictors, each predictor is trained on a small amount of datapoints compared to the traditional approaches.
	Also, each predictor must infer the rotation of all ranges without any anchor angle or any prior information.
	In this situation, the generated training dataset obtained by the proposed method is of great help.
	Indeed, compared to the baseline, \textit{MedErr} in particular shows improvements in performance.
	
	We display the viewpoint evaluation results according to the depth interval in Table.~\ref{viewpoint_eval_interval}.
	As objects far away from the camera are projected as a small size on the 2D image plane, in this case viewpoint estimation is also challenging.
	Therefore, for the baseline using only the limited training dataset, the accuracy of the orientation estimation decreases significantly as the object depth increases.
	However, the inference results for a distant object show robust performance as the proposed method compensates for the boundary limitations of the original dataset.
    Meanwhile, multiple observations can be merged in consideration of the uncertainty of the object's pose estimation.
	We apply the uncertainty estimation for particularly challenging viewpoint estimation.
	Using the uncertainty obtained from \eqref{depth_loss_uncertainty} as a weight, an optimization result considering the pose uncertainty is obtained by \eqref{R_average}.
	In Table.~\ref{viewpoint_eval_interval} we show the estimation results using uncertainty evaluated on KITTI split2.
	In all intervals except $50\sim$, the proposed method shows better performance, and \textit{MedErr} also shows less estimation error.

	\section{Conclusion}
	In this paper, we propose a self-supervised learning method for 3D object understanding from 2D images.
	In order to robustly perform 3D object pose estimation that requires complex non-linear regression, a large-scale training data is necessary.
	However, generating a dataset including 3D annotations for 2D images requires human-level annotations that consumes significant time and cost.
	Therefore, we propose a method to improve the performance by generating training data through self annotation from image sequence which can be acquired by simple navigation.
	When using the independently inferred results for image sequence as annotations for training, the resulting performance is easily bounded by the self-performance of the network.
	In other words, self annotations are bounded to the complexity of the original training dataset so that the importance of the new annotated dataset obtained from sequence disappears.
	To overcome this limitation, we exploit the multiple observations of the object in image sequence.
	After performing data association of each object, the observation result is corrected through outlier removal and weighted averaging.
	Accurate labeling is obtained by re-projecting the obtained global landmark map onto each image.
	Through the proposed technique, the network compensates for its weaknesses in single image prediction, and by actively utilizing generated datasets, estimation results that exceed self-performance are derived.

	
		\clearpage

	
	{\small
	\bibliographystyle{ieee_fullname}
	\bibliography{egbib}}

\begin{thebibliography}{10}\itemsep=-1pt

\bibitem{amrani2020self}
Elad Amrani, Rami Ben-Ari, Inbar Shapira, Tal Hakim, and Alex Bronstein.
\newblock Self-supervised object detection and retrieval using unlabeled
  videos.
\newblock In {\em Proceedings of the IEEE/CVF Conference on Computer Vision and
  Pattern Recognition Workshops}, pages 954--955, 2020.

\bibitem{bansal2018zero}
Ankan Bansal, Karan Sikka, Gaurav Sharma, Rama Chellappa, and Ajay Divakaran.
\newblock Zero-shot object detection.
\newblock In {\em Proceedings of the European Conference on Computer Vision
  (ECCV)}, pages 384--400, 2018.

\bibitem{bochkovskiy2020yolov4}
Alexey Bochkovskiy, Chien-Yao Wang, and Hong-Yuan~Mark Liao.
\newblock Yolov4: Optimal speed and accuracy of object detection.
\newblock {\em arXiv preprint arXiv:2004.10934}, 2020.

\bibitem{bowman2017probabilistic}
Sean~L Bowman, Nikolay Atanasov, Kostas Daniilidis, and George~J Pappas.
\newblock Probabilistic data association for semantic slam.
\newblock In {\em 2017 IEEE International Conference on Robotics and Automation
  (ICRA)}, pages 1722--1729. IEEE, 2017.

\bibitem{M3DRPN}
Garrick Brazil and Xiaoming Liu.
\newblock M3d-rpn: Monocular 3d region proposal network for object detection.
\newblock In {\em Proceedings of the IEEE International Conference on Computer
  Vision}, pages 9287--9296, 2019.

\bibitem{chen20153d}
Xiaozhi Chen, Kaustav Kundu, Yukun Zhu, Andrew~G Berneshawi, Huimin Ma, Sanja
  Fidler, and Raquel Urtasun.
\newblock 3d object proposals for accurate object class detection.
\newblock In {\em Advances in Neural Information Processing Systems}, pages
  424--432. Citeseer, 2015.

\bibitem{chen2018domain}
Yuhua Chen, Wen Li, Christos Sakaridis, Dengxin Dai, and Luc Van~Gool.
\newblock Domain adaptive faster r-cnn for object detection in the wild.
\newblock In {\em Proceedings of the IEEE conference on computer vision and
  pattern recognition}, pages 3339--3348, 2018.

\bibitem{cordts2016cityscapes}
Marius Cordts, Mohamed Omran, Sebastian Ramos, Timo Rehfeld, Markus Enzweiler,
  Rodrigo Benenson, Uwe Franke, Stefan Roth, and Bernt Schiele.
\newblock The cityscapes dataset for semantic urban scene understanding.
\newblock In {\em Proceedings of the IEEE conference on computer vision and
  pattern recognition}, pages 3213--3223, 2016.

\bibitem{dai2020sg}
Angela Dai, Christian Diller, and Matthias Nie{\ss}ner.
\newblock Sg-nn: Sparse generative neural networks for self-supervised scene
  completion of rgb-d scans.
\newblock In {\em Proceedings of the IEEE/CVF Conference on Computer Vision and
  Pattern Recognition}, pages 849--858, 2020.

\bibitem{frost2016object}
Duncan~P Frost, Olaf K{\"a}hler, and David~W Murray.
\newblock Object-aware bundle adjustment for correcting monocular scale drift.
\newblock In {\em 2016 IEEE International Conference on Robotics and Automation
  (ICRA)}, pages 4770--4776. IEEE, 2016.

\bibitem{fu2018deep}
Huan Fu, Mingming Gong, Chaohui Wang, Kayhan Batmanghelich, and Dacheng Tao.
\newblock Deep ordinal regression network for monocular depth estimation.
\newblock In {\em Proceedings of the IEEE Conference on Computer Vision and
  Pattern Recognition}, pages 2002--2011, 2018.

\bibitem{ganin2016domain}
Yaroslav Ganin, Evgeniya Ustinova, Hana Ajakan, Pascal Germain, Hugo
  Larochelle, Fran{\c{c}}ois Laviolette, Mario Marchand, and Victor Lempitsky.
\newblock Domain-adversarial training of neural networks.
\newblock {\em The Journal of Machine Learning Research}, 17(1):2096--2030,
  2016.

\bibitem{KITTIdataset}
Andreas Geiger, Philip Lenz, Christoph Stiller, and Raquel Urtasun.
\newblock Vision meets robotics: The kitti dataset.
\newblock {\em The International Journal of Robotics Research},
  32(11):1231--1237, 2013.

\bibitem{gokcce2015vision}
Fatih G{\"o}k{\c{c}}e, G{\"o}kt{\"u}rk {\"U}{\c{c}}oluk, Erol {\c{S}}ahin, and
  Sinan Kalkan.
\newblock Vision-based detection and distance estimation of micro unmanned
  aerial vehicles.
\newblock {\em Sensors}, 15(9):23805--23846, 2015.

\bibitem{gretton2012kernel}
Arthur Gretton, Karsten~M Borgwardt, Malte~J Rasch, Bernhard Sch{\"o}lkopf, and
  Alexander Smola.
\newblock A kernel two-sample test.
\newblock {\em The Journal of Machine Learning Research}, 13(1):723--773, 2012.

\bibitem{hoffman2013efficient}
Judy Hoffman, Erik Rodner, Jeff Donahue, Trevor Darrell, and Kate Saenko.
\newblock Efficient learning of domain-invariant image representations.
\newblock {\em arXiv preprint arXiv:1301.3224}, 2013.

\bibitem{huang2017densely}
Gao Huang, Zhuang Liu, Laurens Van Der~Maaten, and Kilian~Q Weinberger.
\newblock Densely connected convolutional networks.
\newblock In {\em Proceedings of the IEEE conference on computer vision and
  pattern recognition}, pages 4700--4708, 2017.

\bibitem{huo2020learning}
Yuqi Huo, Hongwei Yi, Zhe Wang, Jianping Shi, Zhiwu Lu, Ping Luo, et~al.
\newblock Learning depth-guided convolutions for monocular 3d object detection.
\newblock In {\em 2020 IEEE/CVF Conference on Computer Vision and Pattern
  Recognition Workshops (CVPRW)}, pages 4306--4315. IEEE, 2020.

\bibitem{NEURIPS2019_d0f4dae8}
Jisoo Jeong, Seungeui Lee, Jeesoo Kim, and Nojun Kwak.
\newblock Consistency-based semi-supervised learning for object detection.
\newblock In H. Wallach, H. Larochelle, A. Beygelzimer, F. d\textquotesingle
  Alch\'{e}-Buc, E. Fox, and R. Garnett, editors, {\em Advances in Neural
  Information Processing Systems}, volume~32. Curran Associates, Inc., 2019.

\bibitem{kendall2017uncertainties}
Alex Kendall and Yarin Gal.
\newblock What uncertainties do we need in bayesian deep learning for computer
  vision?
\newblock {\em arXiv preprint arXiv:1703.04977}, 2017.

\bibitem{li2020improving}
Yandong Li, Di Huang, Danfeng Qin, Liqiang Wang, and Boqing Gong.
\newblock Improving object detection with selective self-supervised
  self-training.
\newblock In {\em European Conference on Computer Vision}, pages 589--607.
  Springer, 2020.

\bibitem{retinanet}
Tsung-Yi Lin, Priyal Goyal, Ross Girshick, Kaiming He, and Piotr Doll{\'a}r.
\newblock Focal loss for dense object detection.
\newblock {\em IEEE transactions on pattern analysis and machine intelligence},
  2018.

\bibitem{lin2014microsoft}
Tsung-Yi Lin, Michael Maire, Serge Belongie, James Hays, Pietro Perona, Deva
  Ramanan, Piotr Doll{\'a}r, and C~Lawrence Zitnick.
\newblock Microsoft coco: Common objects in context.
\newblock In {\em European conference on computer vision}, pages 740--755.
  Springer, 2014.

\bibitem{liu2020self}
Songtao Liu, Zeming Li, and Jian Sun.
\newblock Self-emd: Self-supervised object detection without imagenet.
\newblock {\em arXiv preprint arXiv:2011.13677}, 2020.

\bibitem{SSD}
Wei Liu, Dragomir Anguelov, Dumitru Erhan, Christian Szegedy, Scott Reed,
  Cheng-Yang Fu, and Alexander~C Berg.
\newblock Ssd: Single shot multibox detector.
\newblock In {\em European conference on computer vision}, pages 21--37.
  Springer, 2016.

\bibitem{moon2020multi}
JH Moon, Debasmit Das, and CS~George Lee.
\newblock Multi-step online unsupervised domain adaptation.
\newblock In {\em ICASSP 2020-2020 IEEE International Conference on Acoustics,
  Speech and Signal Processing (ICASSP)}, pages 41172--41576. IEEE, 2020.

\bibitem{mousavian20173d}
Arsalan Mousavian, Dragomir Anguelov, John Flynn, and Jana Kosecka.
\newblock 3d bounding box estimation using deep learning and geometry.
\newblock In {\em Proceedings of the IEEE Conference on Computer Vision and
  Pattern Recognition}, pages 7074--7082, 2017.

\bibitem{categorySpecificSLAM}
Parv Parkhiya, Rishabh Khawad, J~Krishna Murthy, Brojeshwar Bhowmick, and
  K~Madhava Krishna.
\newblock Constructing category-specific models for monocular object-slam.
\newblock {\em arXiv preprint arXiv:1802.09292}, 2018.

\bibitem{yolo9000}
Joseph Redmon and Ali Farhadi.
\newblock Yolo9000: better, faster, stronger.
\newblock {\em arXiv preprint}, 2017.

\bibitem{fasterRCNN}
Shaoqing Ren, Kaiming He, Ross Girshick, and Jian Sun.
\newblock Faster r-cnn: Towards real-time object detection with region proposal
  networks.
\newblock In {\em Advances in neural information processing systems}, pages
  91--99, 2015.

\bibitem{saenko2010adapting}
Kate Saenko, Brian Kulis, Mario Fritz, and Trevor Darrell.
\newblock Adapting visual category models to new domains.
\newblock In {\em European conference on computer vision}, pages 213--226.
  Springer, 2010.

\bibitem{slam++}
Renato~F Salas-Moreno, Richard~A Newcombe, Hauke Strasdat, Paul~HJ Kelly, and
  Andrew~J Davison.
\newblock Slam++: Simultaneous localisation and mapping at the level of
  objects.
\newblock In {\em Proceedings of the IEEE conference on computer vision and
  pattern recognition}, pages 1352--1359, 2013.

\bibitem{simonyan2014very}
Karen Simonyan and Andrew Zisserman.
\newblock Very deep convolutional networks for large-scale image recognition.
\newblock {\em arXiv preprint arXiv:1409.1556}, 2014.

\bibitem{sun2016return}
Baochen Sun, Jiashi Feng, and Kate Saenko.
\newblock Return of frustratingly easy domain adaptation.
\newblock In {\em Thirtieth AAAI Conference on Artificial Intelligence}, 2016.

\bibitem{tekin2018real}
Bugra Tekin, Sudipta~N Sinha, and Pascal Fua.
\newblock Real-time seamless single shot 6d object pose prediction.
\newblock In {\em Proceedings of the IEEE Conference on Computer Vision and
  Pattern Recognition}, pages 292--301, 2018.

\bibitem{tulsiani2015viewpoints}
Shubham Tulsiani and Jitendra Malik.
\newblock Viewpoints and keypoints.
\newblock In {\em Proceedings of the IEEE Conference on Computer Vision and
  Pattern Recognition}, pages 1510--1519, 2015.

\bibitem{tuohy2010distance}
Shane Tuohy, Diarmaid O'Cualain, Edward Jones, and Martin Glavin.
\newblock Distance determination for an automobile environment using inverse
  perspective mapping in opencv.
\newblock 2010.

\bibitem{wu2020eao}
Yanmin Wu, Yunzhou Zhang, Delong Zhu, Yonghui Feng, Sonya Coleman, and Dermot
  Kerr.
\newblock Eao-slam: Monocular semi-dense object slam based on ensemble data
  association.
\newblock {\em arXiv preprint arXiv:2004.12730}, 2020.

\bibitem{xiang2017subcategory}
Yu Xiang, Wongun Choi, Yuanqing Lin, and Silvio Savarese.
\newblock Subcategory-aware convolutional neural networks for object proposals
  and detection.
\newblock In {\em 2017 IEEE winter conference on applications of computer
  vision (WACV)}, pages 924--933. IEEE, 2017.

\bibitem{xiang2014beyond}
Yu Xiang, Roozbeh Mottaghi, and Silvio Savarese.
\newblock Beyond pascal: A benchmark for 3d object detection in the wild.
\newblock In {\em IEEE winter conference on applications of computer vision},
  pages 75--82. IEEE, 2014.

\bibitem{yang2019cubeslam}
Shichao Yang and Sebastian Scherer.
\newblock Cubeslam: Monocular 3-d object slam.
\newblock {\em IEEE Transactions on Robotics}, 35(4):925--938, 2019.

\bibitem{yang2019monocular}
Shichao Yang and Sebastian Scherer.
\newblock Monocular object and plane slam in structured environments.
\newblock {\em IEEE Robotics and Automation Letters}, 4(4):3145--3152, 2019.

\bibitem{yu2018variational}
HW Yu and Beom~Hee Lee.
\newblock A variational feature encoding method of 3d object for probabilistic
  semantic slam.
\newblock In {\em 2018 IEEE/RSJ International Conference on Intelligent Robots
  and Systems (IROS)}, pages 3605--3612. IEEE, 2018.

\bibitem{yu2019variational}
HW Yu, JY Moon, and BH Lee.
\newblock A variational observation model of 3d object for probabilistic
  semantic slam.
\newblock In {\em 2019 International Conference on Robotics and Automation
  (ICRA)}, pages 5866--5872. IEEE, 2019.

\bibitem{yu2021anchor}
Hyeonwoo Yu and Jean Oh.
\newblock Anchor distance for 3d multi-object distance estimation from 2d
  single shot.
\newblock {\em IEEE Robotics and Automation Letters}, 2021.

\bibitem{zhang2020regional}
Yufeng Zhang, Yuxi Li, Mingbi Zhao, and Xiaoyuan Yu.
\newblock A regional regression network for monocular object distance
  estimation.
\newblock In {\em 2020 IEEE International Conference on Multimedia \& Expo
  Workshops (ICMEW)}, pages 1--6. IEEE, 2020.

\bibitem{zhao2020sess}
Na Zhao, Tat-Seng Chua, and Gim~Hee Lee.
\newblock Sess: Self-ensembling semi-supervised 3d object detection.
\newblock In {\em Proceedings of the IEEE/CVF Conference on Computer Vision and
  Pattern Recognition}, pages 11079--11087, 2020.

\bibitem{zhong2018detect}
Fangwei Zhong, Sheng Wang, Ziqi Zhang, and Yizhou Wang.
\newblock Detect-slam: Making object detection and slam mutually beneficial.
\newblock In {\em 2018 IEEE Winter Conference on Applications of Computer
  Vision (WACV)}, pages 1001--1010. IEEE, 2018.

\bibitem{zhu2019learning}
Jing Zhu and Yi Fang.
\newblock Learning object-specific distance from a monocular image.
\newblock In {\em Proceedings of the IEEE International Conference on Computer
  Vision}, pages 3839--3848, 2019.

\end{thebibliography}
	
\end{document}